\documentclass[runningheads]{llncs}

\usepackage[utf8]{inputenc}
\usepackage{amsmath, amssymb} 
\usepackage{graphicx}         
\usepackage{booktabs}         
\usepackage{multirow}         
\usepackage{xcolor}           
\usepackage{cite}             
\usepackage{float}            
\usepackage{url}
\usepackage{algorithm}
\usepackage{algpseudocode}
\usepackage{hyperref}
\usepackage{tikz}
\usetikzlibrary{positioning, arrows.meta, shapes.geometric}

\title{Long-Context Modeling via GSS-Transformer Hybrid Architecture with Learnable Mixing}

\titlerunning{Long-Context Hybrid Modeling}

\author{
  Kuzey Torlak\inst{1} \and
  Hüseyin Arda Arslan\inst{2} \and
  Anıl Dervişoğlu\inst{3} \and
  Beyza Nur Deniz\inst{4} \and
  Onur Boyar\inst{5}
}

\authorrunning{K. Torlak et al.}

\institute{
  Kadıköy Anadolu High School \and
  Politecnico di Torino \and
  Istanbul Technical University \and
  Boğaziçi University \and
  IBM Research - Tokyo
}
\begin{document}

\maketitle
\setcounter{footnote}{0}
\begin{abstract}
Modeling long-range dependencies remains a central challenge in natural language processing. Transformer architectures achieve strong performance via self-attention but scale quadratically ($O(N^2)$) with sequence length, while State Space Models (SSMs) scale linearly ($O(N)$) but suffer from a selective recall bottleneck, struggling to retrieve precise information from compressed states. This creates a fundamental tradeoff between efficiency and perplexity. To tackle these challenges, we propose the \textit{Parallel Hybrid Architecture (PHA)}, which runs Gated State Spaces (GSS), Grouped Query Attention (GQA), and Feed-Forward Networks (FFNs) as independent parallel branches fused by a learnable mixing mechanism. Instead of forcing SSMs to approximate attention or serializing the two paradigms, PHA allows each branch to specialize: GSS captures global context, while attention performs selective retrieval, with FFN providing complementary processing. On WikiText-103, PHA achieves 16.51 PPL at 125M parameters, outperforming Hedgehog (16.70) and H3-125M (23.70). Scaling to 180M parameters yields 16.42 PPL, which gives comparable results with the pure attention baseline while delivering 24\% higher throughput and up to 40\% lower memory usage at long contexts. On OpenWebText, our 125M model achieves 19.72 PPL, outperforming standard Transformers (20.60) and GSS hybrid baselines (19.80). These results demonstrate that separating sequence modeling paradigms into parallel specialists enables Transformer-level perplexity with substantially improved efficiency for long-context language modeling.

\end{abstract}

\section{Introduction}

Most Large Language Models (LLMs) are built on the Transformer architecture \cite{vaswani2017}. Its core mechanism, Self-Attention, allows for modeling dense dependencies between any two tokens in a sequence, regardless of distance. This property makes Transformers what we define as \textit{Retrieval Experts}, which possess the capability to look back and remember specific information (e.g., a name or a number) from the distant past. However, this capability comes at a quadratic cost ($O(N^2)$) in both compute and memory, creating a severe bottleneck for applications requiring long context windows, such as document summarization, genomic modeling, or extended dialogue \cite{transformer_limits}.

To circumvent these limitations, Structured State Space Models (SSMs) like S4 \cite{s4models} and Mamba \cite{gu2023mamba} have emerged. By compressing history into a fixed-size recurrent latent state, SSMs achieve linear scaling ($O(N)$) during training via parallel scans and constant-time inference ($O(1)$) per token. Despite these efficiency gains, pure SSMs often exhibit a recall bottleneck. Since the entire history must be compressed into a state vector of fixed dimension, information loss is inevitable as the sequence length grows. Consequently, SSMs often struggle with needle-in-a-haystack tasks, e.g. retrieving a specific token from thousands of steps ago, where attention-based models thrive.

This trade-off has driven the development of Hybrid Architectures. A prominent example is H3 \cite{h3}, which attempts to mitigate the recall problem by heavily engineering the SSM layer itself to approximate the projection operations of Linear Attention. While effective, this approach introduces an architectural complexity, complicating the layer design to mimic a mechanism that standard Attention already performs naturally. Other approaches, like Jamba \cite{jamba2024}, employ serial stacking (interleaving SSM and Transformer layers), which can still create information bottlenecks where the SSM layers compress the signal before it reaches the attention layers. 

To address these challenges, we propose an architecture that combines an SSM and an attention mechanism without forcing either component to mimic the other. Instead of arranging them sequentially, the two modules operate in parallel, allowing each to specialize: the SSM maintains long-range syntactic coherence through efficient state propagation, while attention performs sparse, high-fidelity retrieval of relevant tokens. We instantiate this idea in the \textit{Parallel Hybrid Architecture (PHA)} (Fig. \ref{fig:architecture}), which integrates Gated State Spaces (GSS) \cite{nguyen2024gss}, Grouped Query Attention (GQA) \cite{gqa}, and a Feed-Forward Network (FFN) through a Learnable Static Mixing layer that determines the relative contribution of each branch to the block output. We primarily validate our approach on the WikiText-103 benchmark and further extend the evaluation to the large-scale OpenWebText corpus to assess its scalability and performance relative to pure SSMs, standard Transformers, and other contemporary hybrid architectures. Using these two datasets enables evaluation of the proposed architecture under different levels of difficulty and data scale. Our results show that the proposed method consistently outperforms or ranks among the top-performing approaches across both datasets.

\section{Background and Related Work}

In this section, we first introduce the evaluation metric and task setting used in our study to provide context for the subsequent discussion. We then organize prior work into three categories: Efficient Transformers, State Space Models, and Hybrid Approaches, and discuss the baselines used in our experimental comparisons.

\subsection{Evaluation Metric}
The fundamental task of Causal Language Modeling is to predict the probability of the next token $x_{t+1}$ given the entire history $x_{0}, \dots, x_t$. The standard metric for evaluating this capability is Perplexity (PPL). Perplexity is defined as the exponentiated average negative log-likelihood of a sequence:
\begin{equation}
    \text{PPL}(X) = \exp \left( -\frac{1}{T} \sum_{t=1}^{T} \log P(x_t \mid x_{<t}) \right)
\end{equation}
Intuitively, PPL measures how surprised the model is by the text; a lower score indicates better prediction and stronger grasp of context. To rigorously test long-range dependency modeling, we utilize the WikiText-103 benchmark \cite{merity2016pointer}. Unlike sentence-level datasets, WikiText-103 consists of full Wikipedia articles, requiring the model to track dependencies across paragraphs and thousands of tokens to achieve a low perplexity score.

\subsection{Architectural Approaches}

Standard Transformers excel as retrieval experts but suffer from the $O(N^2)$ bottleneck. To mitigate this, recurrent variants were introduced. Transformer-XL \cite{transformerxl2019} employs segment-level recurrence, allowing information to propagate beyond fixed windows (achieving $\sim$18.3 PPL in WikiText-103 dataset)\footnote{All PPL values shown in parentheses in this section are reported results on the WikiText-103 dataset from the respective papers.}. Compressive Transformers \cite{compressive2019} improve this by compressing past activations into a memory bank (17.1 PPL). Other optimizations like ALiBi \cite{alibi} and Adaptive Input Representations \cite{adaptiveinput} focus on positional encoding and embedding efficiency, yet they do not fundamentally alter the quadratic attention cost. Another major line of research attempts to approximate the Softmax attention map to achieve linear complexity. Linear Transformers \cite{linear_trans} use kernel feature maps to express attention as an RNN. While fast, they often suffer from poor precision (31.10 PPL). CosFormer \cite{cosformer} attempts to fix this with cosine-based reweighting (23.10 PPL). Recently, Hedgehog \cite{hedgehog} introduced a learnable linear attention that significantly narrows the gap with standard attention (16.70 PPL), representing the current state-of-the-art in this category. Sparse approaches propose an alternative to linear attention by limiting the number of attended tokens. The Routing Transformer \cite{routing} uses k-means clustering to dynamically select relevant tokens, achieving excellent results (15.80 PPL). However, such methods often require complex, hardware-unfriendly implementations (e.g., gather/scatter operations) compared to dense matrix multiplications. SSMs, on the other hand, treat sequence modeling as a continuous signal processing problem. S4 \cite{s4models} uses the Hippo matrix to handle long dependencies efficiently. Although S4 is effective for modeling long-range dependencies, pure S4 (20.95 PPL) struggles with token-level copying behavior that is important for language modeling, where Transformers typically perform better. Other non-attention baselines like AWD-QRNN \cite{awdqrnn} (33.00 PPL) and convolutional models like Hyena \cite{hyena} (18.60 PPL) offer alternative $O(N \log N)$ paths but typically trail behind Transformers in retrieval-heavy benchmarks.

Recognizing that neither pure Attention nor pure SSMs are perfect, hybrid models have gained traction. Early approaches like H3 \cite{h3} attempted to solve the SSM recall weakness by engineering the layer to mimic linear attention (23.7 PPL at 125M). More recently, Hymba \cite{hymba} proposed a parallel architecture similar to ours, employing concurrent SSM and Attention branches. However, Hymba relies on complex, specialized hybrid heads to manage feature mixing. In the following section, we show that our approach simplifies this process by using learnable scalar weights.

\section{Methodology}

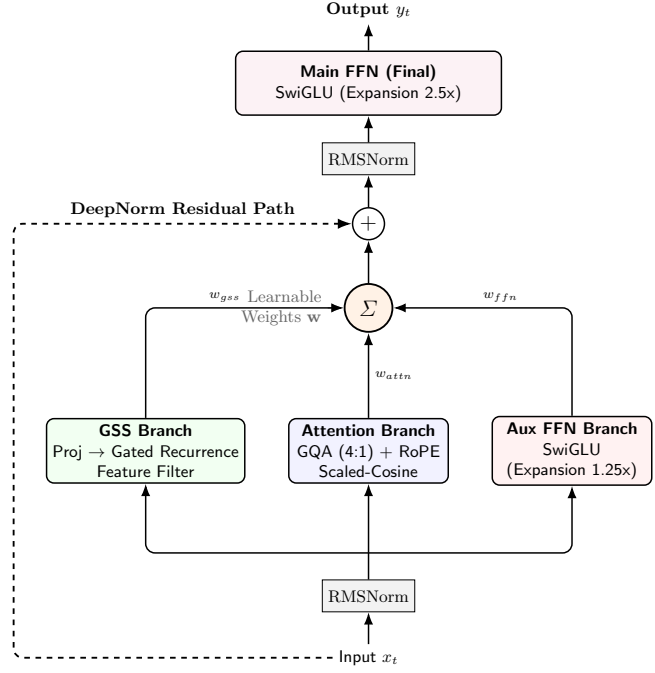
\begin{figure}[H] 
    \centering
    \resizebox{0.7\linewidth}{!}{%
    \begin{tikzpicture}[
        node distance=0.9cm, 
        >=Latex,
        font=\sffamily,
        block/.style={draw, rectangle, rounded corners, minimum height=1.2cm, minimum width=3.0cm, align=center, fill=white, line width=0.8pt},
        smallblock/.style={draw, rectangle, rounded corners=3pt, minimum height=0.8cm, minimum width=2.2cm, align=center, font=\footnotesize, fill=white, line width=0.6pt},
        norm/.style={draw, rectangle, minimum height=0.6cm, minimum width=1.5cm, fill=gray!10, font=\small},
        op/.style={draw, circle, minimum size=0.6cm, fill=white, inner sep=0pt, line width=0.8pt},
        mix/.style={draw, circle, minimum size=0.9cm, fill=orange!10, line width=1pt, font=\large},
        connector/.style={->, thick, rounded corners=5pt}
    ]

    \node (input) {Input $x_t$};
    \node [norm, above=0.6cm of input] (rms1) {RMSNorm};
    \draw [connector] (input) -- (rms1);

    \coordinate [above=0.5cm of rms1] (split);
    \draw [thick] (rms1) -- (split);

    \node [block, above=1.8cm of rms1, fill=blue!5, name=attn] {\textbf{Attention Branch}\\\footnotesize GQA (4:1) + RoPE\\\footnotesize Scaled-Cosine};
    \draw [connector] (split) -- (attn.south);

    \node [block, left=0.8cm of attn, fill=green!5, name=gss] {\textbf{GSS Branch}\\\footnotesize Proj $\to$ Gated Recurrence\\\footnotesize Feature Filter};
    \draw [connector] (split) -| (gss.south);

    \node [block, right=0.8cm of attn, fill=red!5, name=ffn_aux] {\textbf{Aux FFN Branch}\\\footnotesize SwiGLU\\(Expansion 1.25x)};
    \draw [connector] (split) -| (ffn_aux.south);

    \node [mix, above=1.6cm of attn] (sum) {$\Sigma$};
    
    \node [left=0.3cm of sum, font=\small, align=right, text=gray!80!black] (weights) {Learnable\\Weights $\mathbf{w}$};
    \draw [dashed, ->, gray, thick] (weights) -- (sum);

    \draw [connector] (gss.north) |- (sum) node[pos=0.7, above, font=\scriptsize] {$w_{gss}$};
    \draw [connector] (attn.north) -- (sum) node[midway, right, font=\scriptsize] {$w_{attn}$};
    \draw [connector] (ffn_aux.north) |- (sum) node[pos=0.7, above, font=\scriptsize] {$w_{ffn}$};

    \node [op, above=0.8cm of sum, font=\large] (add) {+};
    \draw [connector] (sum) -- (add);
    
    \draw [connector, dashed, line width=1pt] (input.west) -- ++(-6.0,0) coordinate(res_corner) -- (res_corner |- add.west) -- (add.west) node[pos=0.5, above, font=\small] {\textbf{DeepNorm Residual Path}};

    \node [norm, above=0.6cm of add] (rms2) {RMSNorm};
    \draw [connector] (add) -- (rms2);
    
    \node [block, above=0.5cm of rms2, minimum width=5.0cm, fill=purple!5] (final_ffn) {\textbf{Main FFN (Final)}\\\footnotesize SwiGLU (Expansion 2.5x)};
    \draw [connector] (rms2) -- (final_ffn);
    
    \node [above=0.5cm of final_ffn, font=\bfseries] (output) {Output $y_t$};
    \draw [connector] (final_ffn) -- (output);

    \end{tikzpicture}
    }
    \caption{In PHA, the input is processed by three parallel branches: GSS, Attention, and an Auxiliary FFN. These are fused via Learnable Mixing ($\Sigma$). The aggregated signal is then processed by a Main FFN at the end of the block. This structure balances parallel feature extraction with sequential depth processing.}
    \label{fig:architecture}
\end{figure}

We propose a modular \textit{Parallel Hybrid Architecture (PHA)} that processes information through multiple complementary mechanisms operating in parallel. An overview of the architecture is shown in Fig.~\ref{fig:architecture}. The model consists of a stack of $L$ identical layers. Given an input sequence $x \in \mathbb{R}^{T \times D}$, we first embed tokens into a $D$-dimensional space with tied input--output embeddings (i.e., the language model head shares weights with the embedding layer) and scale by $\sqrt{D}$, apply RMSNorm \cite{rmsnorm}, and feed the normalized representation into three parallel branches.

To capture long-range dependencies while maintaining stable state updates, we use a GSS layer. Compared with earlier SSM variants such as S4 or Mamba-style models, GSS provides improved training stability in our parallel setting through its gating mechanism, which regulates how much information is incorporated into the recurrent state. The recurrence is defined as

\begin{equation}
h_t = \sigma(W_g x_t) \odot (A h_{t-1} + B x_t),
\end{equation}

where $W_g \in \mathbb{R}^{H \times D}$ is the gate projection matrix and $H$ denotes the state dimension. The matrix $A$ represents the structured state transition operator, typically initialized using HiPPO-based methods \cite{hippo} to capture long-range memory, while $B$ is the input projection matrix. The sigmoid gate $\sigma(\cdot)$ modulates the state update, allowing the model to control how much new information is incorporated into the hidden state $h_t$. In practice, this mechanism helps maintain a smooth global context representation, while finer token-level interactions are handled by the attention branch.

To capture precise token-level dependencies, we include an attention branch implemented with GQA. The query-to-key/value ratio varies by model scale (see Table~\ref{tab:model_configs}): the 90M and 125M variants use a 4:1 ratio ($H_q = 8$, $H_{kv} = 2$), while the 180M variant uses a 3:1 ratio ($H_q = 12$, $H_{kv} = 4$). In the larger OpenWebText experiments, we use a 4:1 ratio. Compared with standard Multi-Head Attention, this configuration reduces the KV-cache memory footprint by approximately $3\times$ while preserving sufficient representational capacity for complex retrieval patterns.

Positional information is encoded using Rotary Positional Embeddings (RoPE) \cite{rope} with base frequency $\theta = 10{,}000$, enabling robust relative position modeling. To further improve numerical stability in deep hybrid stacks, we replace standard dot-product attention with Scaled-Cosine Attention \cite{swinv2}, which bounds the attention logits and stabilizes training.

Finally, each layer includes a FFN with SwiGLU activation \cite{swiglu}. To balance parameter efficiency and expressiveness, we adopt a Dual-FFN design consisting of a primary processing branch with an expansion factor of $2.5\times$ (or $2.0\times$ for larger datasets) and a secondary auxiliary branch with a smaller expansion factor.

\subsection{Fusion Strategy: Learnable Static Mixing}

The presence of multiple parallel branches requires a mechanism to combine their outputs into a single representation. A straightforward solution would be fixed averaging, but this implicitly assumes that all branches contribute equally at every layer. More complex routing strategies, such as Mixture-of-Experts \cite{moe}, introduce sparsity and load-balancing challenges that are unnecessary in our setting.

Instead, we introduce \textit{Learnable Static Mixing}. We define a set of learnable scalar parameters $p \in \mathbb{R}^3$ corresponding to the Attention, GSS, and FFN branches. The normalized mixing weights $w$ are computed using Softplus to ensure positivity:
\begin{equation}
    w_i = \frac{\text{Softplus}(p_i)}{\sum_{j \in \{attn, ssm, ffn\}} \text{Softplus}(p_j)}.
\end{equation}

Before mixing, each branch output is independently normalized via RMSNorm to ensure comparable scales across branches. The fused output $y_{mix}$ is then obtained as the weighted combination of the normalized branch outputs:
\begin{equation}
    y_{mix} = \sum_{k \in \{attn, ssm, ffn\}} w_k \cdot \text{RMSNorm}(z_k).
\end{equation}

This formulation allows the model to learn the relative importance of each branch through gradient descent, enabling it to emphasize GSS for global context modeling and attention for selective retrieval.

\vspace{-1.0em}

\subsection{Stabilization: DeepNorm Residual Scaling}
Training hybrid models can be unstable due to the interaction between recurrent state updates and attention dynamics. To stabilize optimization, we adopt DeepNorm-style residual scaling \cite{deepnet}, which scales residual connections by $\alpha = (2L)^{0.25}$ to bound the magnitude of layer updates. This design enables stable training of deep networks without relying on complex warm-up schedules or gradient clipping, which are often required in SSM-based models.

\section{Experimental Setup}

To validate the efficacy of our Parallel Hybrid architecture, we conducted rigorous experiments across two datasets, three model scales, and multiple random seeds.

\subsection{Datasets and Tokenization}
In our first sets of experiments, we evaluate our models on the WikiText-103 benchmark \cite{merity2016pointer}. Text is tokenized using the standard GPT-2 Byte-Pair Encoding (BPE) tokenizer \cite{bpe} with a vocabulary of $V=50{,}257$. The corpus is segmented into non-overlapping sequences with a context length of $T=1024$.

In addition, to assess scalability in a larger and more diverse setting, we additionally train on the OpenWebText corpus \cite{openwebtext}, which contains web documents collected from URLs linked to Reddit. Following the H3 protocol, we perform document-level splits with 0.5\% reserved for validation and 0.5\% for testing.

\subsection{Model Configurations}
We evaluate PHA at three parameter scales to study scaling behavior. Table~\ref{tab:model_configs} summarizes the configurations. In this table, FFN Expansion denotes the factor by which the hidden layer expands the input dimension.

\begin{table}[H]
\centering
\caption{\textbf{Model Configurations.} Architectural details for the three PHA variants evaluated.}
\label{tab:model_configs}
\begin{tabular}{l c c c c c c}
\hline
\textbf{Variant} & \textbf{Params} & \textbf{$d_{model}$} & \textbf{Layers} & \textbf{$H_q$} & \textbf{$H_{kv}$} & \textbf{FFN Expansion} \\
\hline
PHA-90M  & 89.7M   & 480 & 14 & 8   & 2 & 2.0$\times$ / 0.67$\times$ \\
PHA-125M & 125M    & 608 & 12 & 8   & 2 & 2.0$\times$ / 0.67$\times$ \\
PHA-180M & 180.5M & 768 & 10 & 12 & 4 & 2.5$\times$ / 1.25$\times$ \\
\hline
\end{tabular}
\end{table}
\vspace{-1.0em}

\subsection{Training Protocol}

All models were trained on a single NVIDIA A100 GPU using PyTorch~\cite{paszke2019pytorch} with full precision (fp32) via the Hugging Face Accelerate library~\cite{accelerate}.

\paragraph{WikiText-103.}
Models were trained for 12 epochs using Cross-Entropy loss with the AdamW optimizer. Initial learning rates were set per scale: $4\times10^{-4}$ for PHA-180M, $4.5\times10^{-4}$ for PHA-125M, and $5\times10^{-4}$ for PHA-90M. We used a ReduceLROnPlateau scheduler (factor $0.75$, patience $3$). The global batch size was $24$. Regularization included dropout $0.1$, weight decay $0.1$, and gradient clipping at norm $1.0$. We applied stochastic weight averaging (SWA) \cite{swa} by tracking the top-5 checkpoints by validation perplexity and averaging the best three at the end of training. All results are reported as mean $\pm$ std across 5 random seeds with deterministic CuDNN settings.

\paragraph{OpenWebText.}
For larger-scale experiments, we trained a $\sim$125M parameter model for 1 epoch ($d_{model}=608$, $n_{layers}=12$, $n_{heads}=8$, GQA $4{:}1$). FFN expansion factors were reduced to $2\times$ and $0.67\times$. We used a cosine learning rate schedule with $2{,}000$ warmup steps, peak LR $4.5\times10^{-4}$, and decay to $10\%$ of the maximum. The batch size was $24$. SWA was applied using the \emph{Top-5} checkpoint group.

\begin{figure}[H]
    \centering
    \includegraphics[width=0.75\linewidth]{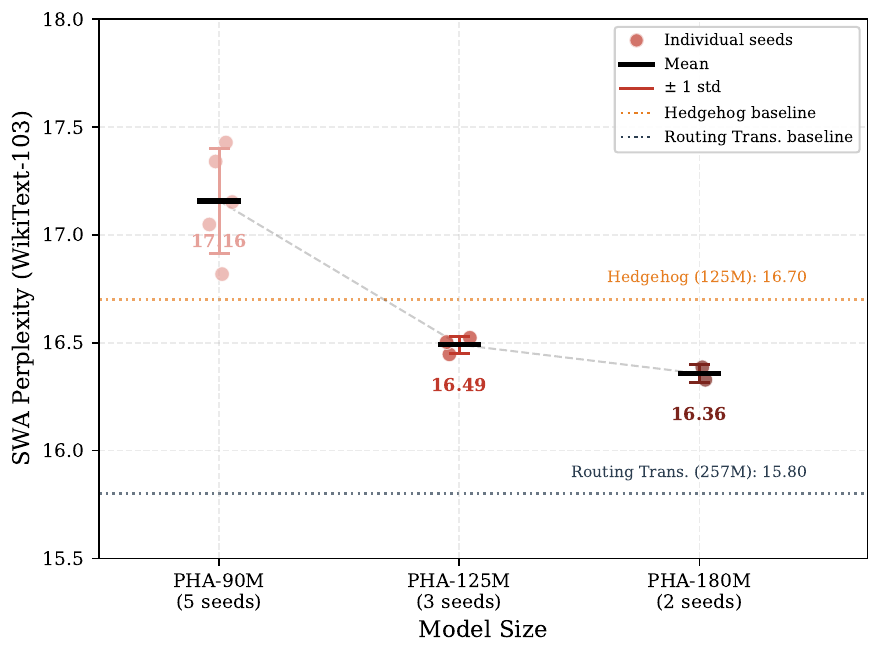}
    \caption{Individual seed results (circles), means (horizontal bars), and $\pm$1 std error bars for each PHA model variants. The 125M and 180M variants consistently outperform Hedgehog (125M), approaching Routing Transformer (257M) quality at a fraction of the parameter count.}
    \label{fig:pha_seeds}
\end{figure}

\section{Results and Analysis}

In this section, we present our experimental results on WikiText-103 and OpenWebText. We then perform an extensive set of ablation studies to evaluate the contribution of individual components of the proposed parallel hybrid architecture.

\subsection{WikiText-103 Experiments}
We benchmark our models against the suite of architectures discussed in Section 2. For our models, we report the mean $\pm$ std across seeds. Table \ref{tab:125m_comparison} benchmarks our 125M model. From this table, we observe that our methodology provides better results than various benchmark studies, including another hybrid methodology H3-Hybrid \cite{h3} by a large margin.
\begin{table}[H]
\centering
\caption{\textbf{Comparison of $\sim$125M Parameter Models on WikiText-103.} Our PHA-125M outperforms all direct competitors.}
\label{tab:125m_comparison}
\resizebox{\textwidth}{!}{%
\begin{tabular}{l c c c c c}
\hline
\textbf{Model} & \textbf{Params} & \textbf{Context} & \textbf{Tokenizer} & \textbf{Test PPL} ($\downarrow$) & \textbf{Arch. Type} \\
\hline
\textbf{Ours (PHA-125M)}$^\dagger$ & \textbf{125M} & \textbf{1024} & \textbf{GPT-2 BPE} & \textbf{16.51 $\pm$ 0.12} & \textbf{Hybrid (GSS+Attn)} \\
Hedgehog \cite{hedgehog} & 125M & 1024 & GPT-2 BPE & 16.70 & Linear Attn \\
Hyena-3-slim \cite{hyena} & 125M & 2048 & GPT-2 BPE & 18.50 & Convolutional \\
Hyena-3 \cite{hyena} & 125M & 2048 & GPT-2 BPE & 18.60 & Convolutional \\
H3-Hybrid (125M) \cite{h3} & 125M & 2048 & GPT-2 BPE & 23.70 & Hybrid SSM \\
Linear Transformer \cite{linear_trans} & 90M & 384 & Word-level* & 31.10 & Linear Attn \\
AWD-QRNN \cite{awdqrnn} & 185M & - & Word-level* & 33.00 & RNN \\
GPT-2 Small \cite{radford2019} & 117M & 1024 & BPE & 37.50 & Transformer \\
BERT-CAS \cite{bertcas} & 140M & 512 & WordPiece & 39.85 & Encoder-Decoder \\
\hline
\end{tabular}
}
\end{table}
\vspace{-1.0em}

We also compared our approach against models with more than 200M parameters. Table \ref{tab:large_comparison} compares our 180M model against significantly larger baselines. Notably, many of these baselines use Word-level tokenizers, which typically yield numerically lower perplexity than BPE (as the vocabulary is smaller). Despite this disadvantage, our BPE-based model remains highly competitive.

\begin{table}[H]
\centering
\caption{\textbf{Comparison with Larger Baselines (200M+).} Despite having fewer parameters (180M), our model outperforms major baselines like S4, Transformer-XL, and Compressive Transformer.}
\label{tab:large_comparison}
\resizebox{\textwidth}{!}{%
\begin{tabular}{l c c c c c}
\hline
\textbf{Model} & \textbf{Params} & \textbf{Context} & \textbf{Tokenizer} & \textbf{Test PPL} ($\downarrow$) & \textbf{Arch. Type} \\
\hline
Routing Transformer \cite{routing} & 257M & 4096 & Word-level* & \textbf{15.80} & Sparse Attn \\
\textbf{Ours (PHA-180M)}$^\dagger$ & \textbf{180M} & \textbf{1024} & \textbf{GPT-2 BPE} & \textbf{16.42 $\pm$ 0.09} & \textbf{Hybrid} \\
H3-355M \cite{h3} & 355M & 2048 & GPT-2 BPE & 16.90 & Hybrid SSM \\
Compressive Transformer \cite{compressive2019} & 257M & 2048 & Word-level* & 17.10 & Recurrent \\
Transformer-XL \cite{transformerxl2019} & 257M & 1600 & Word-level* & 18.30 & Recurrent \\
ALiBi Transformer \cite{alibi} & 247M & 3072 & Word-level* & 18.30 & Transformer \\
Adaptive Input Rep. \cite{adaptiveinput} & 247M & 2560 & Word-level* & 18.70 & Transformer \\
BERT-Large-CAS \cite{bertcas} & 395M & 512 & WordPiece & 20.42 & Transformer \\
S4 \cite{s4models} & 249M & 8192 & GPT-2 BPE & 20.95 & Pure SSM \\
CosFormer \cite{cosformer} & 200M+ & 512 & Word-level* & 23.10 & Linear Attn \\
Standard Softmax Trans. \cite{cosformer} & 477M & 512 & Word-level* & 24.92 & Transformer \\
\hline
\end{tabular}
}
\end{table}
\vspace{-1.0em}

\begin{figure}[htbp]
    \centering
    \includegraphics[width=0.85\textwidth]{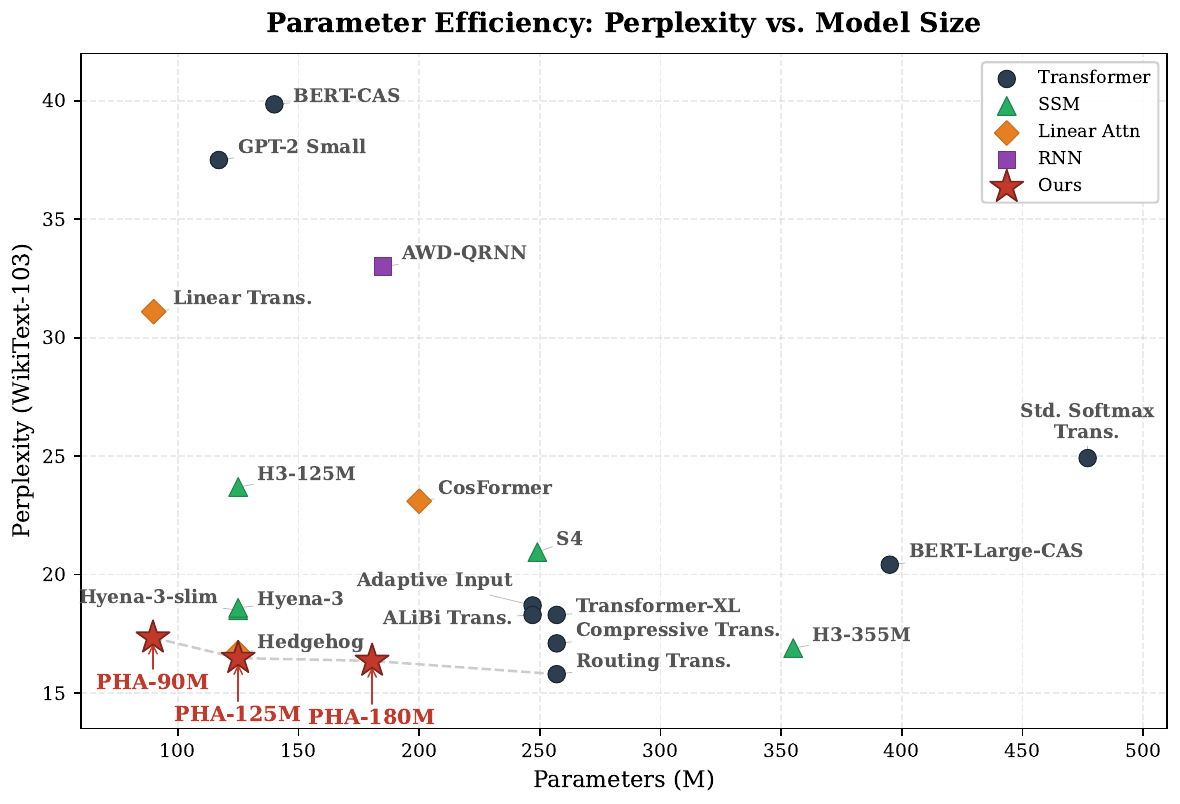}
    \caption{\textbf{Parameter Efficiency: Perplexity vs.\ Model Size.} All PHA variants (stars) lie on the Pareto frontier of best-performing models. The dashed line traces the best models across scales. PHA-125M matches Hedgehog at 1.3$\times$ fewer perplexity points while PHA-180M approaches Routing Transformer quality at 70\% of the parameter count.}
    \label{fig:pareto}
\end{figure}

In addition to above experiments, to demonstrate the model's scaling capability across different model sizes, along with 125M and 180M parameter models, we also experimented with 90M parameter model and tested them in WikiText-103 dataset. Figure \ref{fig:pha_seeds} shows the scaling behaviour of the proposed PHA and benchmarks it against 125M Hedgehog model, while Table~\ref{tab:scaling} presents PPL values along with the accuracy. Results show that PHA performance increases as the number of parameters increase. In addition, Fig.~\ref{fig:pareto} compares perplexity on WikiText-103 with model size across several approaches in the literature. The results show that PHA achieves competitive performance while using fewer parameters than many existing methods.

\begin{table}[H]
\centering
\caption{\textbf{Scaling Behavior of PHA.} Perplexity improves consistently from 90M to 180M. The 125M$\rightarrow$180M jump ($+$55M params) reduces PPL by only 0.09, suggesting diminishing returns and that the 125M variant offers the best quality-per-parameter ratio at this training budget.}
\label{tab:scaling}
\begin{tabular}{l c c c c}
\hline
\textbf{Model} & \textbf{Params} & \textbf{Config} & \textbf{SWA PPL} ($\downarrow$) & \textbf{Accuracy (\%)} \\
\hline
PHA-90M$^\dagger$  & 89.7M   & 480d $\times$ 14L & 17.16 $\pm$ 0.24 & 45.95 $\pm$ 0.21 \\
PHA-125M$^\dagger$ & 125M    & 608d $\times$ 12L & 16.51 $\pm$ 0.12 & 46.62 $\pm$ 0.12 \\
PHA-180M$^\dagger$ & 180.5M & 768d $\times$ 10L & 16.42 $\pm$ 0.09 & 46.78 $\pm$ 0.13 \\
\hline
\end{tabular}
\end{table}
\vspace{-1.0em}

\subsection{OpenWebText Experiments}
To validate our architecture in a large-data regime, we evaluated the 125M variant on the OpenWebText corpus following the H3 methodology \cite{h3}. All baseline values are sourced from \cite{h3}, where every model (Transformer, GSS, and H3 variants) was trained for 100K steps with a batch size of 512 and sequence length 1024, accumulating approximately 50 billion tokens. Our model, by contrast, was trained for a single epoch with a batch size of 24, seeing approximately 12 billion tokens in total, roughly 23\% of the baseline token budget.

Despite this significantly reduced training budget, PHA-125M achieves a perplexity of 19.72 using SWA over the Top-5 checkpoints, outperforming the standard Transformer (20.60) and the GSS Hybrid (19.80), while remaining competitive with the H3 Hybrid (19.60). Table~\ref{tab:openwebtext} summarizes these results. The fact that our architecture reaches competitive performance with less than a quarter of the training tokens suggests that the parallel hybrid design offers improved sample efficiency relative to both pure and serial hybrid baselines.

\begin{table}[htbp]
\centering
\caption{\textbf{OpenWebText Evaluation (125M Class).} All baselines were trained on ${\sim}$50B tokens; our model was trained on ${\sim}$12B tokens (1 epoch). Despite seeing roughly 4$\times$ fewer tokens, PHA-125M outperforms pure SSMs and the standard Transformer.}
\label{tab:openwebtext}
\begin{tabular}{l c @{\hspace{1.5em}} c @{\hspace{1.5em}} c}
\toprule
\textbf{Model} & \textbf{Params} & \textbf{Tokens Seen} & \textbf{Test PPL} ($\downarrow$) \\
\midrule
GSS \cite{nguyen2024gss} & $\sim$125M & $\sim$50B & 24.00 \\
H3 \cite{h3} & $\sim$125M & $\sim$50B & 21.00 \\
Transformer & $\sim$125M & $\sim$50B & 20.60 \\
GSS Hybrid \cite{nguyen2024gss} & $\sim$125M & $\sim$50B & 19.80 \\
H3 Hybrid (125M) \cite{h3} & $\sim$125M & $\sim$50B & {19.60} \\
\textbf{Ours (PHA-125M)} & \textbf{125M} & \textbf{$\sim$12B} & \textbf{19.72} \\
\bottomrule
\end{tabular}
\end{table}
\vspace{-1.0em}

Overall, results on WikiText-103 and OpenWebText demonstrate that PHA achieves competitive or superior performance across both benchmarks. On OpenWebText in particular, the model's strong showing at a fraction of the training compute suggests that the parallel hybrid design extracts more useful signal per token than conventional architectures, making it a promising direction for compute-constrained settings.

\subsection{Ablation Studies}
\label{sec:ablations}

We conduct a set of controlled ablations on WikiText-103 to isolate three questions: 
(i) whether GSS and Attention should be composed in parallel or sequentially, 
(ii) how much each branch contributes to language modeling performance, and 
(iii) why the full hybrid remains preferable even when Attention-Only achieves similar perplexity. 
Unless otherwise noted, all variants are trained with matched hyperparameters at the $\sim$180M scale.

\subsubsection{Composition Strategy: Parallel vs.\ Sequential}
\label{sec:parallel_vs_sequential}

A central design choice in PHA is to run GSS and Attention as independent parallel branches with learnable mixing, rather than stacking them in a fixed serial order. To test whether this composition itself matters, we construct a Sequential Hybrid baseline using the same components---GSS, GQA with the same head configuration, SwiGLU FFN, DeepNorm, RoPE, and Scaled-Cosine Attention---but arranged in a strict serial pipeline:
\begin{equation}
    x \;\xrightarrow{\text{Norm}}\; \text{Attn} \;\xrightarrow{+\text{residual}}\; \text{Norm} \;\xrightarrow{\text{GSS}} \;\xrightarrow{+\text{residual}}\; \text{Norm} \;\xrightarrow{\text{FFN}} \;\xrightarrow{+\text{residual}}\; x'.
\end{equation}
In this design, GSS only sees attention-transformed representations, FFN only sees GSS-transformed features, and the relative contribution of branches is fixed by the ordering. To match the parameter budget, the sequential variant uses 12 layers (vs.\ 10 for PHA), totaling 185.8M parameters.

Table~\ref{tab:parallel_vs_sequential} shows that the parallel design performs slightly better despite being smaller: PHA-180M achieves lower SWA perplexity (16.42 vs.\ 16.46) with 5.3M fewer parameters, while matching the sequential model in best-checkpoint perplexity. Although the gap is modest, it is notable because the sequential baseline is deeper, which typically favors serial architectures.

\begin{table}[H]
\centering
\caption{\textbf{Parallel vs. Sequential Composition on WikiText-103.} Both models use identical components (GSS, GQA, SwiGLU, DeepNorm) at a comparable parameter budget. The parallel design achieves lower perplexity with fewer parameters, supporting the use of architectural independence. $^\dagger$Mean $\pm$ std across seeds.}
\label{tab:parallel_vs_sequential}
\resizebox{\textwidth}{!}{%
\begin{tabular}{l c c c c c}
\hline
\textbf{Composition} & \textbf{Params} & \textbf{Config} & \textbf{SWA PPL} ($\downarrow$) & \textbf{Best-Ckpt PPL} & \textbf{Acc (\%)} \\
\hline
\textbf{Parallel (PHA-180M)}$^\dagger$ & \textbf{180.5M} & 768d $\times$ 10L & \textbf{16.42 $\pm$ 0.09} & \textbf{16.95} & \textbf{46.78} \\
Sequential & 185.8M & 768d $\times$ 12L & 16.46 & 16.95 & 46.93 \\
\hline
\end{tabular}
}
\end{table}
\vspace{-1.0em}

The result suggests that the benefit comes not just from the component set, but from how the components interact. In PHA, all branches receive the same normalized input and can specialize independently, while the learned mixer adapts their relative importance by layer. In contrast, the sequential design constrains information flow through a fixed pipeline.

\subsubsection{Branch Contribution}
\label{sec:branch_ablation}

Having established that parallel composition is beneficial, we next ask which branches are most important within that parallel structure. Starting from the full PHA-180M model, we train variants that remove one or more branches while keeping the overall budget comparable. We also include an Attention-Only baseline as the natural endpoint obtained by removing both GSS and the auxiliary FFN.

\begin{table}[H]
\centering
\caption{\textbf{Branch Removal Ablation (PHA-180M on WikiText-103).} Each row removes one or more branches from the full architecture while maintaining a comparable parameter budget. $\Delta$ PPL denotes the gap relative to the full model. Removing Attention causes by far the largest degradation, while removing GSS or the auxiliary FFN yields more modest losses. $^\dagger$Mean $\pm$ std across seeds.}
\label{tab:branch_ablation}
\resizebox{\textwidth}{!}{%
\begin{tabular}{l l c c l}
\hline
\textbf{Configuration} & \textbf{Branches Kept} & \textbf{SWA PPL} ($\downarrow$) & \textbf{$\Delta$ PPL} & \textbf{Removed} \\
\hline
Attention-Only (Transformer) & Attention              & 16.32 & --0.10 & GSS + Aux FFN \\
\textbf{Full PHA-180M}$^\dagger$ & GSS + Attention + FFN  & \textbf{16.42 $\pm$ 0.09} & --- & --- \\
Attention + FFN (--GSS)      & Attention + Aux FFN    & 16.82 & +0.40 & GSS \\
Attention + GSS (--FFN)      & Attention + GSS        & 17.12 & +0.70 & Aux FFN \\
GSS + FFN (--Attention)      & GSS + Aux FFN          & 20.69 & +4.27 & Attention \\
GSS Only (--Attn --FFN)      & GSS                    & 21.30 & +4.88 & Attention + Aux FFN \\
\hline
\end{tabular}
}
\end{table}
\vspace{-1.0em}

These extensive analyses and ablation studies clarify the role of each design choice in PHA. Parallel composition is preferable to sequential stacking even when using identical components, indicating that architectural independence itself is beneficial. Attention provides the critical retrieval capability required for competitive perplexity, while GSS and the auxiliary FFN improve the computational tradeoff around this capability.

At the same time, the small gap between the full model and Attention-Only (16.42 vs.\ 16.32) suggests an important nuance: at this scale, Attention alone is sufficient for near-optimal perplexity. This naturally raises the next question: if Attention-Only is so strong, why use the hybrid at all?


The answer is efficiency. Although PHA and the Attention-Only baseline achieve near-identical perplexity, the hybrid is substantially more efficient at longer contexts because it offloads a large fraction of sequence processing to the linear-time GSS branch. Table~\ref{tab:efficiency_scaling} shows that the gap widens with sequence length: at 8192 tokens, PHA is 23.9\% faster, while also using 22.0\% less memory. The memory benefit is largest at shorter contexts, reaching 39.6\% at 2048 tokens.

\begin{table}[H]
\centering
\caption{\textbf{Efficiency: PHA-180M vs. Attention-Only Transformer.} Throughput and memory comparison across sequence lengths. Both models achieve near-identical perplexity, but PHA is consistently more efficient, especially at longer contexts.}
\label{tab:efficiency_scaling}
\begin{tabular}{l | c c | c c | c c}
\hline
\multirow{2}{*}{\textbf{Seq Length}} & \multicolumn{2}{c|}{\textbf{Throughput (tok/s)} $\uparrow$} & \multicolumn{2}{c|}{\textbf{Memory (GB)} $\downarrow$} & \multicolumn{2}{c}{\textbf{Hybrid Gain}} \\
 & \textbf{Attn-Only} & \textbf{Hybrid} & \textbf{Attn-Only} & \textbf{Hybrid} & \textbf{Speed} & \textbf{Mem.} \\
\hline
2048  & 32,620 & \textbf{34,409} & 2.12 & \textbf{1.28} & +5.5\%  & \textbf{--39.6\%} \\
4096  & 31,382 & \textbf{35,276} & 2.70 & \textbf{1.86} & +12.4\% & --31.1\% \\
8192  & 26,683 & \textbf{33,058} & 3.87 & \textbf{3.02} & \textbf{+23.9\%} & --22.0\% \\
16384 & N/A     & N/A             & 6.19 & \textbf{5.34} & N/A     & --13.7\% \\
\hline
\end{tabular}

\end{table}
\vspace{-1.0em}

This computational advantage is consistent with the learned mixing behavior in Table~\ref{tab:layer_weights}. The mixing weights show a stable ``Sandwich'' pattern: GSS dominates near the input and output layers, while Attention becomes more prominent in the middle layers. Thus, branch utilization and branch criticality are not the same. Attention is the indispensable retrieval expert, but GSS is the main workhorse that carries much of the routine sequence processing when both are available. This explains why the full hybrid can match Transformer-level perplexity while being more efficient.

\begin{figure}[htbp]
    \centering
    \includegraphics[width=0.95\textwidth]{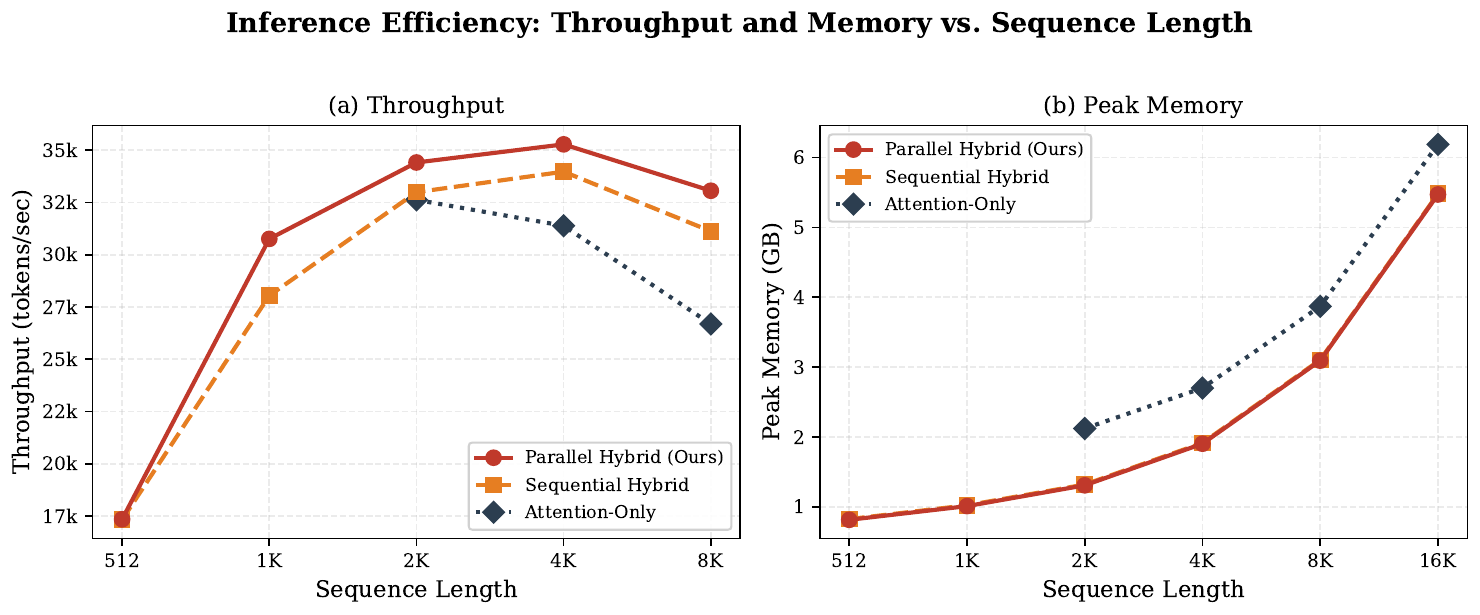}
    \caption{\textbf{Inference Efficiency: Throughput and Memory vs.\ Sequence Length.} The hybrid model maintains higher throughput and lower memory usage than the Attention-Only baseline, with the throughput advantage increasing at longer contexts.}
    \label{fig:efficiency}
\end{figure}

\subsubsection{Learnable Mixing}
\label{sec:mixing_analysis}

Beyond removing individual components, it is also informative to examine how the model allocates computation across branches during training. A natural question is whether the mixer collapses to a single dominant branch or instead distributes weight across branches in a structured way. To investigate this, we analyze the learned mixing weights of PHA-180M across layers and multiple checkpoints (Steps 48k--58k). The distribution remains stable throughout training, indicating that the mixing mechanism does not collapse to a single branch.

Table~\ref{tab:layer_weights} reveals a clear layerwise specialization pattern. GSS receives the largest weight near the input and output layers, while Attention becomes most prominent in the middle layers. This forms a functional ``Sandwich'' structure: GSS dominates the boundaries where global context is established and consolidated, while Attention becomes more active in intermediate layers where selective retrieval from the context is most useful. The auxiliary FFN consistently receives a smaller but non-negligible share, suggesting a supporting processing role rather than serving as the primary information pathway.
\vspace{-1em}
\begin{table}[H]
\centering
\caption{\textbf{Layer-wise Branch Weight Analysis (PHA-180M).} The learned mixer discovers a stable ``Sandwich'' specialization: GSS dominates near the boundaries to anchor global context, while Attention becomes more prominent in the middle layers for selective retrieval. Average across layers: GSS 46.9\%, Attention 31.2\%, FFN 21.9\%.}
\label{tab:layer_weights}
\resizebox{0.9\textwidth}{!}{%
\begin{tabular}{l c c c}
\hline
\textbf{Layer} & \textbf{GSS (Context)} & \textbf{Attention (Retrieval)} & \textbf{FFN (Processing)} \\
\hline
L0  & 54.0\% & 29.1\% & 17.0\% \\
L1  & 53.2\% & 20.0\% & 26.8\% \\
L2  & 52.4\% & 22.7\% & 24.9\% \\
L3  & 43.9\% & 33.1\% & 23.1\% \\
L4  & 39.9\% & \textbf{41.4\%} & 18.7\% \\
L5  & 42.5\% & 37.7\% & 19.9\% \\
L6  & 39.1\% & \textbf{41.9\%} & 19.1\% \\
L7  & 41.8\% & 37.9\% & 20.3\% \\
L8  & 47.6\% & 29.7\% & 22.7\% \\
L9  & \textbf{54.6\%} & 18.9\% & 26.5\% \\
\hline
\textbf{Average} & \textbf{46.9\%} & \textbf{31.2\%} & \textbf{21.9\%} \\
\hline
\end{tabular}
}
\end{table}
\vspace{-1.0em}

This analysis also helps interpret the branch-removal results. Although GSS receives the largest average weight, \emph{utilization} and \emph{necessity} are distinct properties: Attention provides the indispensable retrieval capability, while GSS acts as the primary computational workhorse when Attention is present.

Taken together, the learned mixing behavior supports the central design intuition of PHA: the branches specialize across depth and cooperate in a structured way, rather than operating redundantly. Parallel composition is preferable to sequential stacking even with identical components, Attention provides the critical retrieval capability, and GSS improves the computational tradeoff around it. As a result, PHA achieves Transformer-level perplexity while replacing a substantial portion of quadratic computation with linear-time processing.

\section{Conclusion}

We introduced the PHA, which combines Gated State Spaces and Grouped Query Attention through a learnable static mixing mechanism to address long-context modeling.

Experiments on WikiText-103 across multiple model scales show that PHA achieves strong performance in both quality and efficiency. Our 125M model surpasses the H3-125M baseline by over 7 perplexity points, while the 180M model outperforms H3-355M despite using roughly half the parameters. At the same time, PHA matches the quality of pure attention models while delivering higher throughput and lower memory usage at long context lengths.

Controlled ablations confirm that the parallel design is superior to sequential hybridization, improving both perplexity and runtime efficiency. Further analysis reveals complementary roles between the branches: attention provides critical token-level retrieval, while the state-space component carries the majority of the computation for global context modeling.

Finally, consistent performance across multiple random seeds and model scales demonstrates that the proposed architecture is robust and scalable. Additional experiments on OpenWebText further validate these findings, where the 125M model achieves competitive perplexity while maintaining the efficiency advantages of the hybrid design. Overall, our results suggest that combining specialized sequence modeling mechanisms in parallel offers a practical and scalable approach for efficient long-context language modeling.


\end{document}